\documentclass[letterpaper, 10 pt, conference]{ieeeconf}
\IEEEoverridecommandlockouts                              %

\usepackage{graphicx} %
\usepackage{amsmath} %
\usepackage{amssymb}  %
\usepackage{float}
\usepackage[ruled,linesnumbered]{algorithm2e}

\SetCommentSty{mycommfont}
\usepackage{tabularx}
\usepackage{bbm}
\usepackage{dsfont}
\usepackage{booktabs}
\usepackage[sort,compress]{cite}
\usepackage{balance} %

\usepackage{url}

\usepackage{tikz}
\usepackage{textcomp}
\usepackage{hyperref}
\usepackage{subcaption}
\usepackage{ifthen}

\DeclareCaptionLabelSeparator{periodspace}{.\quad}
\newboolean{arxiv}
\setboolean{arxiv}{true}

\def\doi{DOI: \href{https://doi.org/XX.YYYYY/X.YYYYY}{XX.YYYYY/X.YYYYY}}
\newcommand\copyrighttext{%
  \footnotesize \textcopyright \the\year{} IEEE. Personal use of this material is permitted.
  Permission from IEEE must be obtained for all other uses, in any current or future   media, including reprinting/republishing this material for advertising or promotional   purposes, creating new collective works, for resale or redistribution to servers or   lists, or reuse of any copyrighted component of this work in other works.
  }
\newcommand\copyrightnotice{%
\begin{tikzpicture}[remember picture,overlay]
\node[anchor=south,yshift=10pt] at (current page.south) {\fbox{\parbox{\dimexpr\textwidth-\fboxsep-\fboxrule\relax}{\copyrighttext}}};
\end{tikzpicture}%
}

\SetKwProg{myCodeBlock}{}{}{}

\makeatother

\title{\LARGE \bf
Multi-Robot Distributed Semantic Mapping in Unfamiliar Environments \\ 
through Online Matching of Learned Representations
}

\author{Stewart Jamieson$^{1,2}$, Kaveh Fathian$^2$, Kasra Khosoussi$^2$, Jonathan P. How$^2$, Yogesh Girdhar$^3$%
\thanks{$^{1}$S. Jamieson is with the MIT-WHOI Joint Program in Oceanography/Applied Ocean Science and Engineering
        {\tt\small sjamieson@whoi.edu }}%
\thanks{$^{2}$S. Jamieson, K. Fathian, K. Khosoussi, and J. P. How are with the Department of Aeronautics and Astronautics at the Massachusetts Institute of Technology (MIT)
        {\tt\small \{sjamieson, kavehf, kasra, jhow\}@mit.edu }}%
\thanks{$^{3}$Y. Girdhar is with the Applied Ocean Physics and Engineering Department at the Woods Hole Oceanographic Institution (WHOI)
        {\tt\small yogi@whoi.edu}}%
\thanks{*This work was supported in part by NSF-NRI Award Number 1734400, by ARL DCIST under Cooperative Agreement Number W911NF-17-2-0181, and by ONR under BRC award N000141712072.}%
}

\begin{document}

\maketitle
\thispagestyle{empty}
\pagestyle{empty}

\begin{abstract}
\ifthenelse{\boolean{arxiv}}{\copyrightnotice}{}%
We present a solution to multi-robot distributed semantic mapping of novel and unfamiliar environments.
Most state-of-the-art semantic mapping systems are based on supervised learning algorithms that cannot classify novel observations online.
While unsupervised learning algorithms can invent labels for novel observations, approaches to detect when multiple robots have independently developed their own labels for the same new class are prone to erroneous or inconsistent matches.
These issues worsen as the number of robots in the system increases and prevent fusing the local maps produced by each robot into a consistent global map, which is crucial for cooperative planning and joint mission summarization.
Our proposed solution overcomes these obstacles by having each robot learn an unsupervised semantic scene model online and use a multiway matching algorithm to identify consistent sets of matches between learned semantic labels belonging to different robots. 
Compared to the state of the art, the proposed solution produces 20-60\% higher quality global maps that do not degrade even as many more local maps are fused.

\end{abstract}

\section*{Open source Software}
An implementation of this solution is contributed in the ``Sunshine'' 3D semantic mapping ROS package, a general purpose single- and multi-robot semantic mapping system: \url{https://gitlab.com/warplab/ros/sunshine}.

\section{Introduction}
\label{sec:intro}

\begin{figure}[t]
    \centering
    \includegraphics[width=\columnwidth]{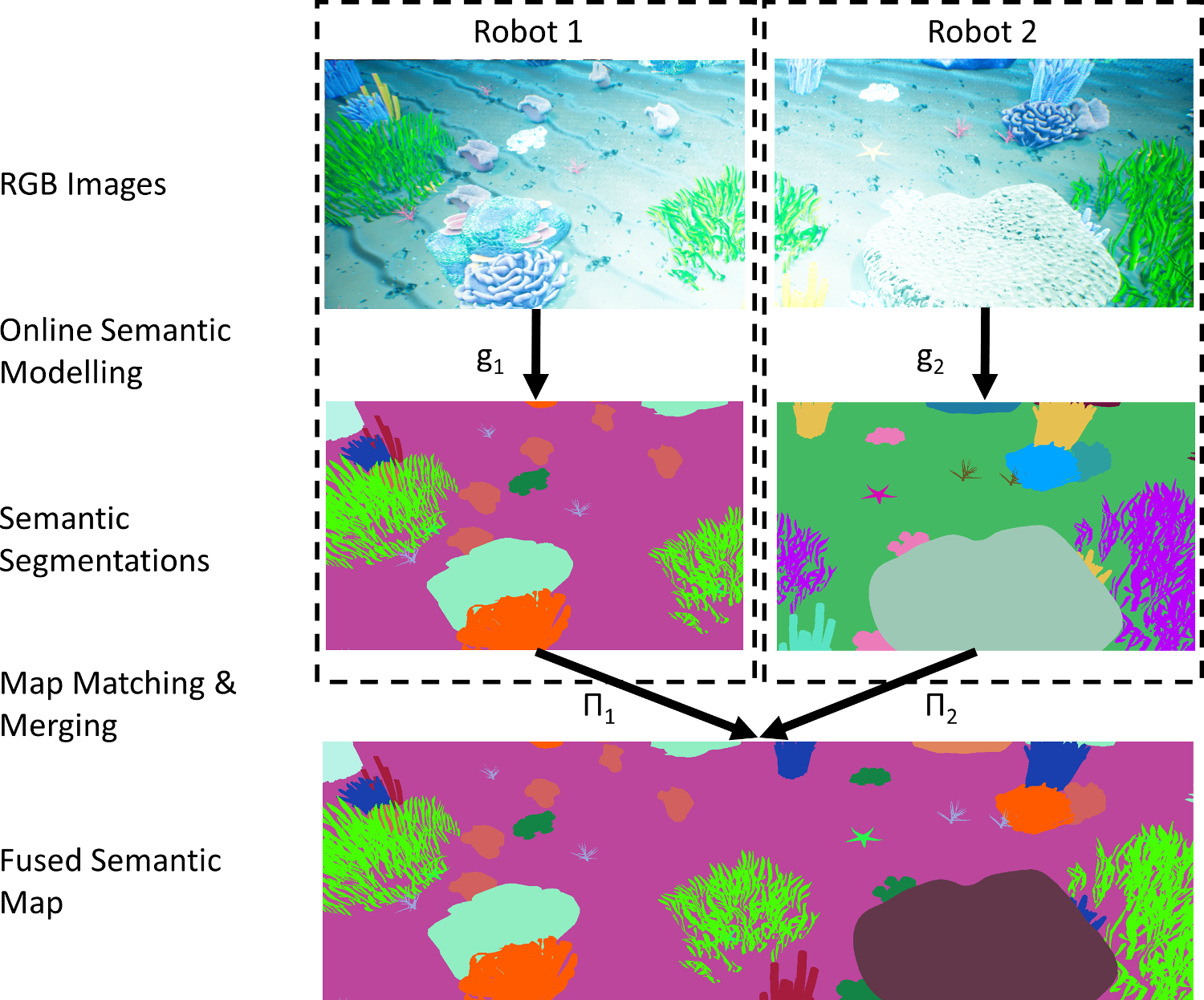}
    \caption{When robots develop different individual semantic models $g_i$, they solve the correspondences $\Pi_i$ into a global (shared) semantic language in order to fuse their results. Images are from environment \#2 (see Fig.~\ref{fig:airsim}).}
    \label{fig:overview}
\end{figure}

Semantic mapping is a relatively young field that was initially motivated by giving robots a spatial awareness of nearby terrains, objects, and activities~\cite{Wolf2008}. Semantic maps describe the world using a set of classes, and have been used with great effectiveness in solving many field robotics problems such as mission summarization~\cite{Flaspohler2017}, object-based SLAM~\cite{Bowman2017,Himri2018,Zhang2019}, and context-aware planning~\cite{Everett2019,Jamieson2020}. Great progress has been made in improving the accuracy of semantic mapping systems by leveraging deep learning models trained on large datasets~\cite{Sunderhauf2017,Pham2019,Rosinol2019,Nakajima2019,Grinvald2019}. However, most of these systems do not support \textit{a priori} unknown classes, which are an essential part of scientific exploration. 
For example, in an underwater exploration scenario an explicit goal is novelty detection, e.g., discovering new species or geological phenomena. This complication makes it crucial to find algorithms that use unsupervised learning to develop new semantic representations \textit{online}, so that the semantic mapping system can detect and classify novel observations.

Another important aspect of scientific exploration is that the task often involves large and unknown environments, which can be very time consuming to cover with a single robot.
Furthermore, many large-scale phenomena of scientific interest, such as mass migrations, feeding events, and geological activities like volcanism, are transient and dynamic and can thus be easily missed or insufficiently covered by a single robot.
These issues necessitate using a team of robots working in parallel for \textit{distributed} mapping and exploration.

When learning semantic representations online using unsupervised algorithms, the learned models are egocentric.
Fig.~\ref{fig:overview} demonstrates this can be an issue: one robot has learned a different permutation of the same semantic labels (represented with colors) learned by the other robot. 
In addition to the unknown permutation between corresponding labels, some labels learned by one robot may not correspond to any label observed by another robot, or 
a single label may represent the union of multiple labels learned by another robot. 
Therefore, multi-robot distributed semantic mapping with learned representations requires correctly estimating the total number of distinct labels, and associating and fusing labels that correspond to the same semantic category.

This work presents a novel system for multi-robot distributed semantic mapping that addresses the previously described issues.
Each robot uses an online semantic 3D mapping system to model its own observations and create a high quality semantic map.
The robots explore the target environment in parallel, sharing their learned semantic maps and models with each other and with the human operator whenever communication constraints permit.
Finally, a multiway matching algorithm that can run on any robot estimates the total number of unique phenomena observed across the robot team, finds matches between the same phenomena labeled differently by various robots, and fuses the local maps to obtain a consistent global map across all robots.

\section{Related Works}
\label{sec:related-works}

Spatiotemporal topic models (STMs)~\cite{Blei2003,Cao2007,Wang2007} are a class of unsupervised learning algorithms that has been specifically augmented for realtime semantic mapping in novel environments~\cite{Girdhar2014a,Girdhar2016}. BNP-ROST is an STM that adaptively develops semantic labels online as new phenomena are observed~\cite{Girdhar2016a}, and has been used for 3D semantic mapping in bandwidth-limited environments without any pre-training~\cite{Girdhar2019_ICRA}. STMs can create high quality semantic maps in realtime by leveraging the spatial and temporal correlations between observations and using efficient sampling algorithms to discover good semantic representations online~\cite{Girdhar2015,Girdhar2019_ICRA}. While there has been progress in other unsupervised semantic image segmentation approaches~\cite{Thoma2016}, including deep-learning based methods~\cite{Xia2017}, by not leveraging these correlations they produce lower quality maps of large environments.

Doherty et al.~\cite{Doherty2018} showed that repeatedly matching the topic models of two robots with the Hungarian algorithm~\cite{Kuhn1955}, merging them together, and distributing the merged model would result in both robots converging to a single set of good semantic labels, even if this was done at a low frequency. To our knowledge,~\cite{Doherty2018} is the only prior work that has explored multi-robot distributed semantic mapping with representations learned online. The present work improves upon~\cite{Doherty2018} with a novel solution for matching many ($N\gg2$) semantic maps that is more robust to major variations across what each robot sees, and does not rely on distributing the merged topic model throughout the entire robot team. Eliminating the need to distribute the merged model halves communication bandwidth usage, and makes the approach more robust to transient connection failures.

Multiway matching algorithms are a class of data association techniques that
leverage the transitivity property (cycle consistency) to rectify wrong correspondences and construct a unified representation from the partial and noisy observations of multiple agents.  
Multiway matching leads to superior accuracy compared to classical pairwise approaches (e.g.,~\cite{Kuhn1955}), however it has combinatorial complexity.
State-of-the-art methods consider approximations of this problem via convex relaxations \cite{Pachauri2013,Chen2014,Hu2018b,Yu2016}, spectral relaxations \cite{Zhou2015,Maset2017}, or graph clustering \cite{Yan2016} to obtain a solution in polynomial time.
In this work we use CLEAR~\cite{Fathian2019}, a spectral clustering based approach, to match topic models because it is one of the most efficient multiway matching algorithms with leading performance in both precision and recall.

\section{Problem Setup}
\label{sec:setup}

Let us denote the environment to be mapped as $E\subset \mathbb{R}^3$. We partition $E$ into a grid of disjoint cells (boxes) $\mathcal{B}=\{b_i\}$ such that $E = \cup_{i} b_i$. An oracle (e.g., the human operator) could assign to each box a distribution over human-defined semantic labels $\mathcal{Z}^H$ that represent its semantic contents. We assume that if these boxes are sufficiently small, each box can be effectively represented by a single dominant label. We thus define the ground truth semantic segmentation $f:\mathcal{B} \rightarrow \mathcal{Z}^H$, where $f(b)\in\mathcal{Z}^H$ represents the human label for $b\in\mathcal{B}$. The model $f$ and set $\mathcal{Z}^H$ are unknown \textit{a priori} because the operator did not know what would be found during the mission, and communication bandwidth limitations prevent the operator from seeing most of the observations until the mission is over and the robots are recovered. The goal of the robot team is to construct a fused semantic map $g : \mathcal{B} \rightarrow \mathcal{Z}^G$, where $\mathcal{Z}^G$ is a shared set of learned semantic labels, such that $g(b)$ is ``similar'' to $f(b)$. A metric to evaluate this similarity will be presented in Section~\ref{sec:methods}.

We assume that the team consists of $N$ autonomous robots. By timestep $t$, the $n^\text{th}$ robot has collected its own set of localized image observations and used them to build, in an unsupervised manner, a local semantic map $g_{n,t}: \mathcal{B} \rightarrow \mathcal{Z}^n_{t}$, where $\mathcal{Z}^n_{t}$ is the set of semantic labels the robot has developed to describe its own observations.\footnote{In practice, we assume $g_{n,t}$ is learned from RGB-D image observations $\{o_{n,\tau}\}_{\tau=1}^t$ paired with estimated camera poses $\{x_{n,\tau}\in\mathtt{SE}(3)\}_{\tau=1}^t$.}
Due to the egocentricity of unsupervised semantic mapping, robots which observe different phenomena, or the same but in a different order, will likely develop disparate semantic models, as in Fig.~\ref{fig:overview}. In order to construct a fused semantic map, these $N$ unique semantic models must first be fused to use a common set of labels.
Therefore, the team must construct a set of global semantic labels $\mathcal{Z}^G_t$ and a set of correspondences $\Pi_t = \{\Pi_{n,t}:\mathcal{Z}^{n}_{t}\rightarrow  \mathcal{Z}^G_{t}\}_{n=1}^N$ that translate individual robots' labels $\mathcal{Z}^{n}_{t}$ into $\mathcal{Z}^G_{t}$.
Given $\Pi_t$, the individual semantic maps can be fused into a single global map $g_t : \mathcal{B} \rightarrow  \mathcal{Z}^G_t$ for time $t$ (see Fig.~\ref{fig:overview}). Each label can be computed as $g_t(b)=\Pi_{n^\star_b,t}(g_{n^\star_b,t}(b))$ where $n^\star_b$ is the index of the robot that most recently visited and observed the cell $b\in\mathcal{B}$.

\begin{figure*}[t]
    \centering
    \includegraphics[width=0.9\textwidth]{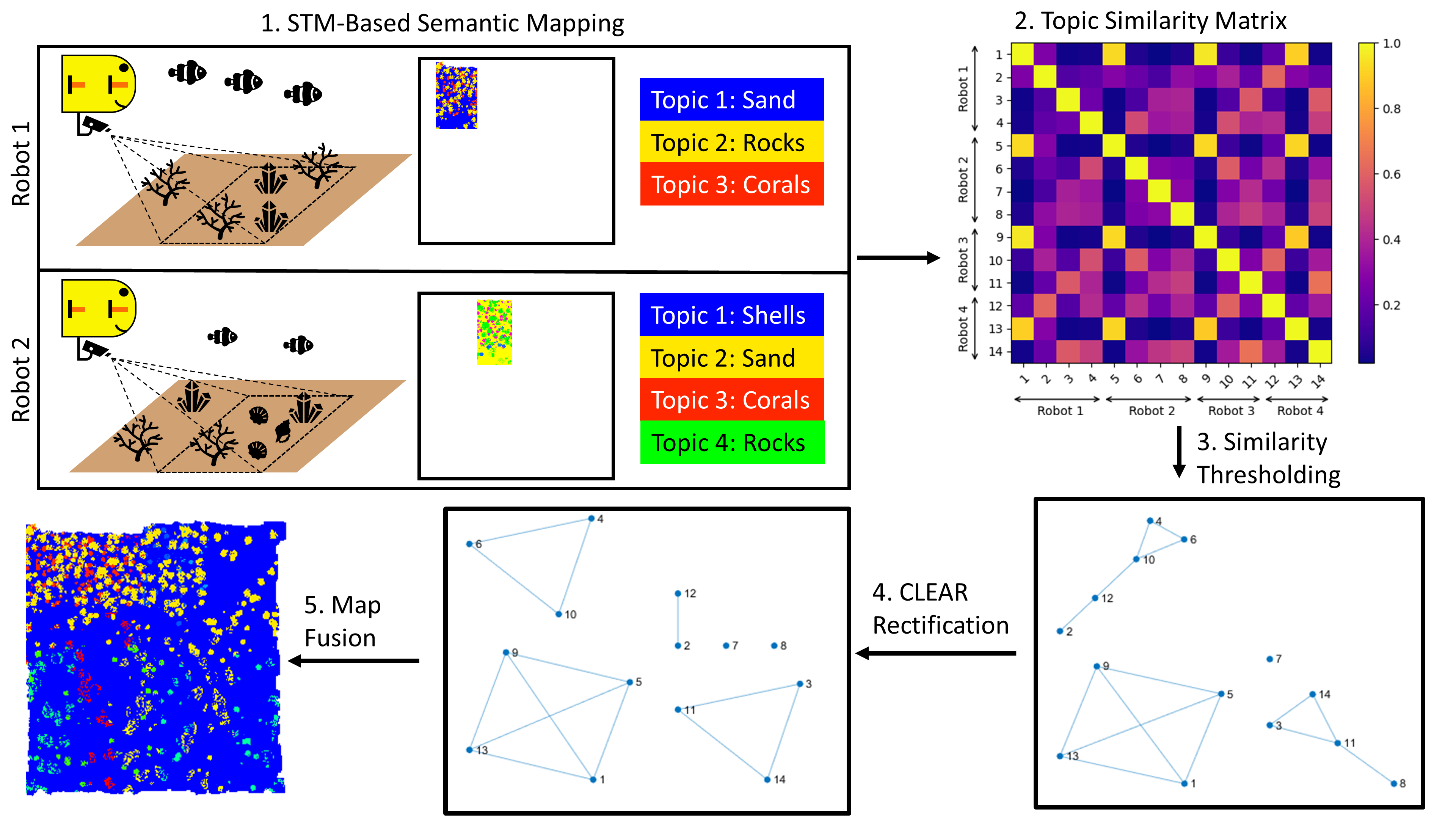}
    \caption{Our proposed system is composed of five stages. Each robot first learns an individual semantic model online which it uses to describe its own observations. When a fused map is required, a topic similarity matrix is constructed by using a similarity metric for topic descriptors to compute all pairwise similarity scores. A noisy association graph is produced by treating each topic as a vertex and using similarities above a specified threshold as edges. The noisy graph is rectified using the CLEAR multiway matching algorithm to produce a consistent topic matching, which has the form of a cluster graph. These topic matches are used to fuse the individual maps into one consistently labelled global map.}
    \label{fig:matching-approach}
\end{figure*}

\section{Consistent Online Topic Matching}
\label{sec:algorithm}

We present an algorithm to produce a fused semantic map from semantic maps built by robots with disparate semantic models. \textit{Consistency} ensures that the fused model created by any agent is the same across all agents, and \textit{online} indicates that our algorithm will be run during the mission whenever a new matching is required. Typically, we expect a human operator would run our algorithm at a central node whenever they require an updated global map, however it is computationally lightweight enough that it may be run by any robot that has collected other robots' semantic maps.

The stages of the proposed system are presented graphically in Fig.~\ref{fig:matching-approach} and in detail by the following subsections.

\subsection{Online STM-Based Semantic Mapping}

The proposed approach has each robot construct a spatiotemporal topic model online as the basis for its individual semantic map, as in~\cite{Girdhar2019_ICRA}. Nonetheless, it would be straightforward to adapt this system to use any similar unsupervised online semantic mapping module. 

While an exhaustive description of the BNP-ROST model used is left to~\cite{Girdhar2016a}, a few details are relevant here. First, the model requires no pre-training, although it can be bootstrapped with topics for common phenomena. Second, the model is tuned by varying the feature vocabulary, the spatiotemporal grid cell size, and three scalar hyperparameters. Typically, the vocabulary is domain-specific while the cell size and the scalar hyperparameters are tuned for a specific mission.\footnote{While they may be tuned with a dataset representative of the target environment, these hyperparameters encode only abstract information about the domain and thus tend to generalize well to novel environments.} All robots on the same mission use the same hyperparameters. Finally, each robot's topic model uses a stochastic process to develop an ever-evolving set of topics online based on its own visual observations. At time $t$ the $n^\text{th}$ robot has $K_{n,t}$ topics, so $\mathcal{Z}^{n}_{t}=\{1,\dots,K_{n,t}\}$. Each topic $z_k\in\mathcal{Z}^{n}_{t}$ is characterized by a semantic ``descriptor'' $\phi_k\in\Delta^V$, a distribution over ``words'' in the predefined vocabulary of size $V$~\cite{Girdhar2014a}, where $\Delta^V = \{p\in\mathbb{R}^V_+ : \Vert p \Vert_1 = 1\}$. Each grid cell $b_i$ of the environment $E$ is labelled with a single maximum likelihood topic, which may change as the model evolves; thus, we will start using the terms ``topic'' and ``label'' interchangeably.

\subsection{Computing Topic Similarity}

We require a similarity metric that measures how similar two topics are to each other in order to identify when multiple robots have developed any semantically equivalent topics.
Since each descriptor $\phi_k$ represents a probability mass function, it is natural to consider similarity metrics that operate on discrete probability distributions. Total Variation Distance (TVD)~\cite{Levin2017} measures the largest possible difference in probability that two topics assign to the same set of words.
Thus, the Topic Overlap (TO),
\begin{equation}
    \text{TO}(\phi_1,\phi_2) = 1-\text{TVD}(\phi_1, \phi_2),
\end{equation}
is a similarity metric that represents the total probability mass which both $\phi_1$ and $\phi_2$ assign similarly. It can be computed using the identity $\text{TVD}(\phi_1, \phi_2) = \frac{1}{2} \Vert \phi_1 - \phi_2 \Vert_1$.

Another metric commonly used for comparing topic descriptors is Cosine Similarity (CS)~\cite{Aletras2014}, which computes the cosine of the angle between two descriptors as
\begin{equation}
    \text{CS}(\phi_1, \phi_2) = \frac{\phi_1 \cdot \phi_2}{\Vert\phi_1\Vert \Vert\phi_2\Vert}.
\end{equation}
The CS metric is about as efficient to compute as TO, but assigns a higher score when $\phi_1$ and $\phi_2$ are very similar and a lower score when they are very dissimilar. This may be preferable because, as the stochastic nature of topic models means that the descriptors $\{\phi_k\}$ fluctuate constantly, two topics with the same semantic meaning are likely to have slightly different descriptors at any given time.

Each similarity metric presented above is symmetric and bounded by $[0,1]$, where a score of $0.0$ indicates two topics have no words in common, and $1.0$ indicates that two topics are exactly the same. We use the chosen similarity metric $s$ to construct the \textit{pairwise similarity graph}, a weighted and undirected graph in which vertices are topics and edge weights are the similarity of the adjoining vertices; it is represented in Fig.~\ref{fig:matching-approach} by the topic similarity matrix.

\subsection{Constructing the Noisy Association Graph}

In practice, similarity metrics are ``noisy'' in that topics which a human would judge to have the same semantic meaning may not have a similarity score of $1.0$, and topics with very different semantic meanings may not have a similarity of $0.0$.
The pairwise similarity graph is simplified by removing edges with weights below some $\sigma\in\left(0,1\right)$ that represents low similarity, and setting weights above $\sigma$ to $1$, resulting in the unweighted \textit{noisy association graph}.

\label{subsec:thresholding}In general, a good threshold $\sigma$ for considering two topics to be ``sufficiently similar'' will depend on factors including the similarity metric $s$, the topic model hyperparameters, and the subjective opinion of the human operator.
It is difficult to choose $\sigma$ analytically because the expected topic growth rate and average inter-topic similarity are complicated functions of the topic model hyperparameters. A simple solution for choosing $\sigma$ is to collect a validation set of topics developed by robots in past missions, for which the human operator can infer their semantic meanings, and then tune $\sigma$ low enough that the algorithm merges as many topics with the equivalent meanings as possible but high enough that it does not match distinct topics. In the training dataset used to choose the topic model hyperparameters, to be described in Section~\ref{sec:methods}, a threshold of $\sigma=0.75$ was found to work well with both topic similarity metrics.
However, even if the threshold $\sigma$ is chosen well, it may not be obvious from this graph how many unique topics should be used in the final map, or which sets of topics would form a consistent matching.

\subsection{Rectifying the Noisy Association Graph}
\label{subsec:clear-rectification}

When visualized as a graph where vertices represent topics and edges represent matches, a consistent matching has the structure of a \textit{cluster graph}. This is a graph composed of disjoint fully-connected components, so that any two topics in the same component are matched and no topics are matched between components. In this cluster graph, the number of distinct topic labels developed by the entire robot team is equal to the number of disjoint components.

CLEAR~\cite{Fathian2019} is a spectral clustering algorithm that estimates the closest cluster graph to a noisy association graph (see Fig.~\ref{fig:matching-approach}).
A key reason for choosing CLEAR is that it is one of the fastest algorithms to perform multiway matching with high accuracy.
The Laplacian $L$ of the noisy association graph is a matrix defined in terms of the graph adjacency matrix $A$ and degree matrix $D$ as $L = D - A$, where
\begin{align}
\left[A\right]_{ij} & =\begin{cases}
1, & s\left(\phi_i,\phi_j\right)\ge\sigma\\
0, & \text{otherwise}
\end{cases}\\
\left[D\right]_{ij} & =\begin{cases}
\sum_{k=1}^{N}\left[A\right]_{ik}, & i=j\\
0, & \text{otherwise}
\end{cases}
\end{align}

CLEAR uses a special normalization of the Laplacian based on the degree matrix plus identity, denoted by $L_\text{nrm}$,  
to identify clusters of semantic labels in the noisy association graph with high pairwise similarity. 
The number of eigenvalues of $L_\text{nrm}$ less than $0.5$ is a robust estimate of the number of global labels, $\vert\mathcal{Z}^G_t\vert$. 
CLEAR then uses the eigenvectors of $L_\text{nrm}$ to find a consistent set of label correspondences $\Pi_t$.

\section{Experimental Methodology}
\label{sec:methods}

\begin{figure}[t]
    \centering
    \subcaptionbox{Environment \# 1.}{\includegraphics[width=.45\columnwidth]{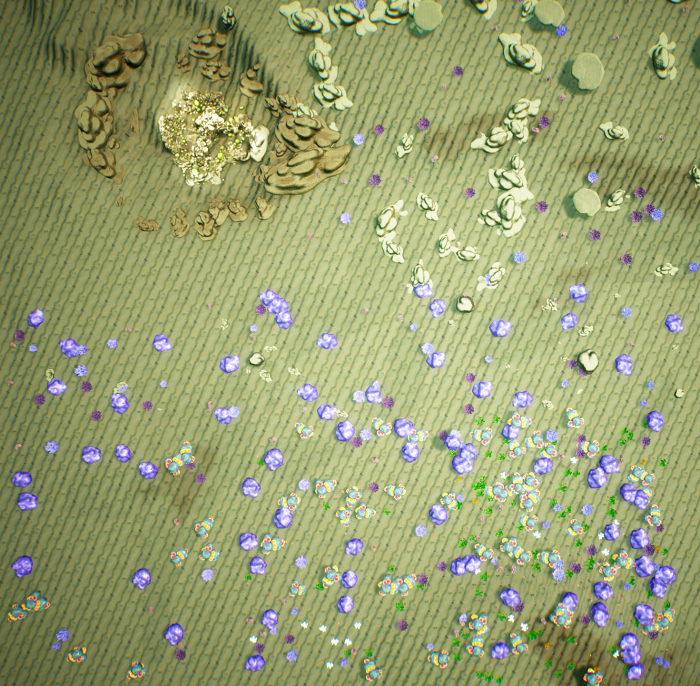}}
    \subcaptionbox{Environment \# 2.}{\includegraphics[width=.45\columnwidth]{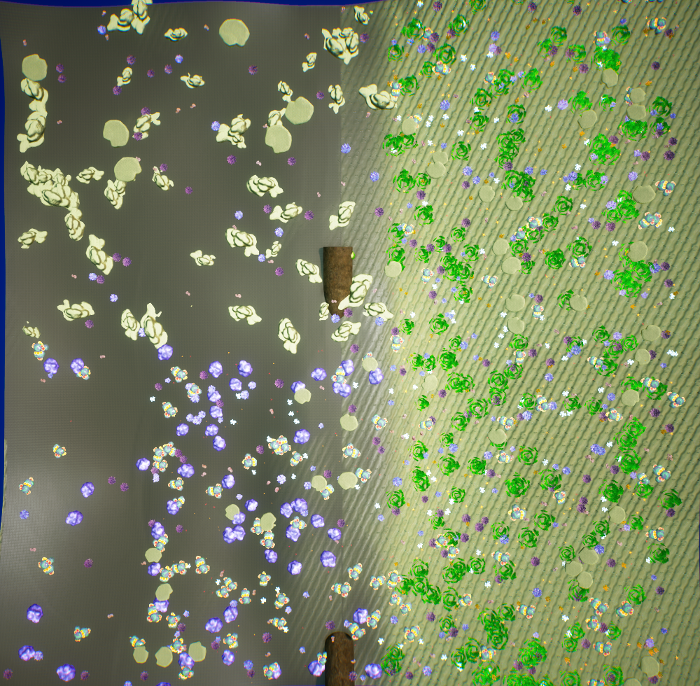}}
    \caption{Top-down views of the two simulated test environments used in the experiments. Each map is approximately 250m$\times$250m, and contains a rich variety of coral species, seaweed, and rocks.}
    \label{fig:airsim}
    \vspace{-.05in}
\end{figure}

\begin{figure*}[t]
    \centering
    \subcaptionbox{In environment \#1, sparser and less varied phenomena (coral species) were spread throughout a uniformly sandy reef. The prevalence of sand in both shaded and well lit conditions tended to cause the single robot to develop two sand topics, reducing its performance.
    \label{fig:gt-ami-1}}{
    \includegraphics[width=0.46\textwidth]{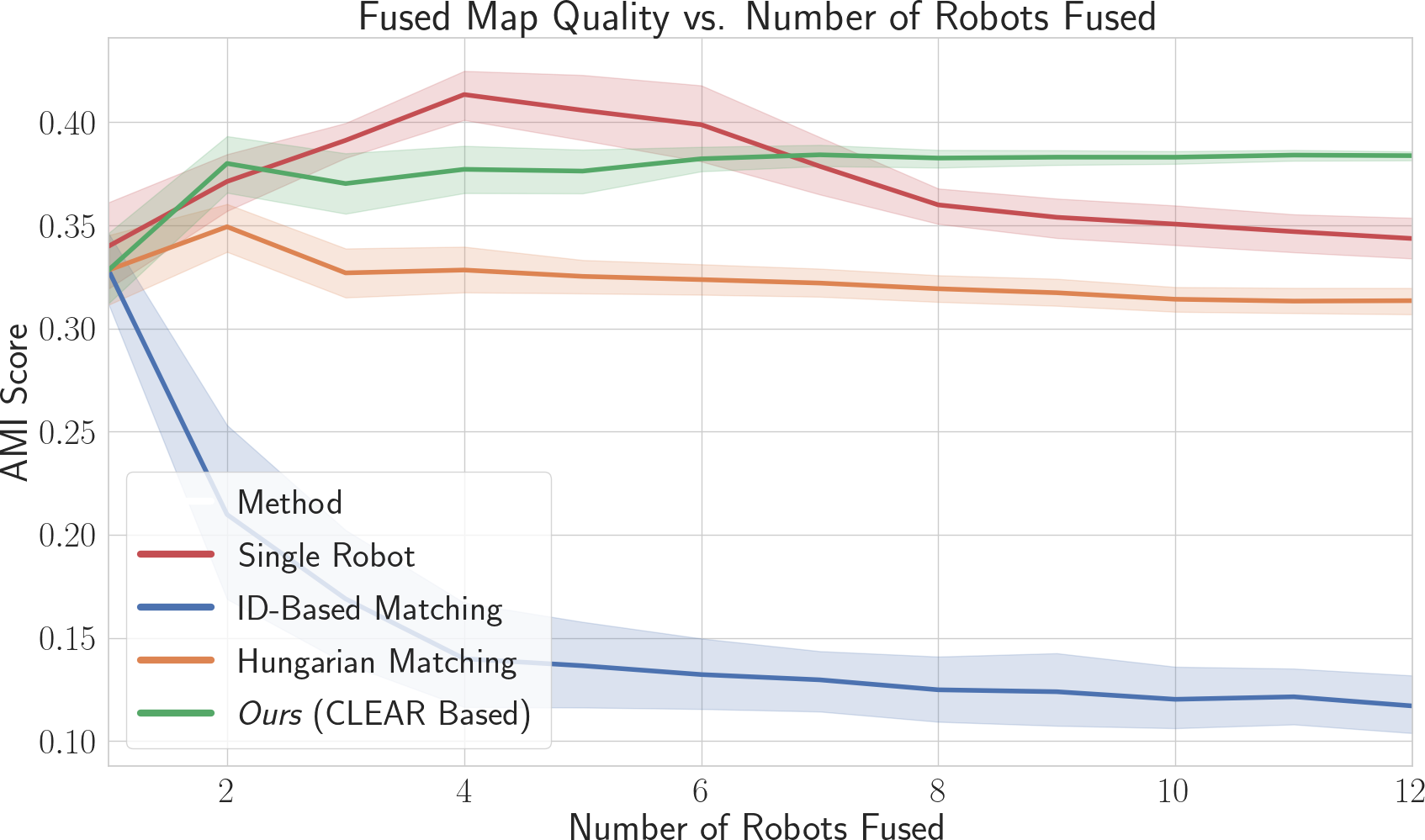}}\hspace{0.05\textwidth}
    \subcaptionbox{In environment \#2, more variation between robots in the phenomena (coral species) and terrains observed meant that the Hungarian algorithm's assumptions were violated, leading to reduced performance compared to the proposed CLEAR-based approach.\label{fig:gt-ami-2}}{
    \includegraphics[width=0.46\textwidth]{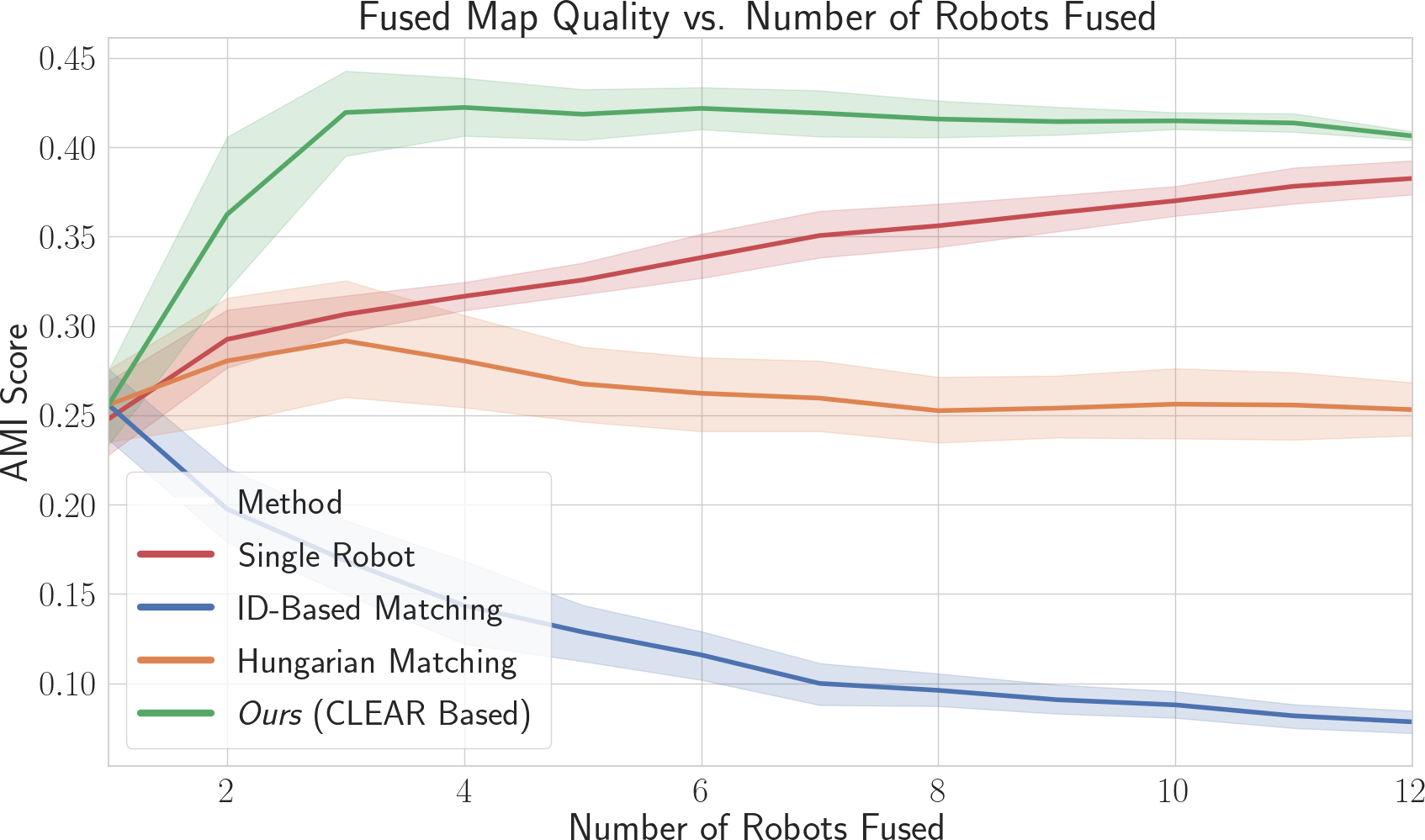}}
    
    \caption{The AMI scores between the fused maps and corresponding ground truth maps extracted from the simulator demonstrate that the multi-robot distributed mapping system with CLEAR-based label matching outperforms all other multi-robot approaches. Error bars represent the 95\% confidence interval of the mean score. The score of a single-robot that explored the same total area as the corresponding fused maps is shown in red, for comparison.}
    \label{fig:gt-ami}
    \vspace{-.1in}
\end{figure*}

The proposed consistent online topic matching system was evaluated using semantic mapping experiments in two unique high-resolution 3D simulated coral reef environments produced in the Unreal Engine~\cite{EpicGames2019} using the Automatic Coral Generator package~\cite{Kelint2016}. A top-down view of each simulated environment is presented in Fig.~\ref{fig:airsim}. In each experiment, a team of 12 simulated robots traversed one of the two environments and each collected 250 RGB-D observations using the AirSim plugin~\cite{Shah2017}. AirSim also provided a ground truth semantic segmentation for each image. Each robot was given noiseless localization information and ran a new release of the BNP-ROST~\cite{Girdhar2016a} spatiotemporal topic model called  ``Sunshine'', which is conceptually identical to the system presented in~\cite{Girdhar2019_ICRA} but redesigned for ease of use and with optimized code to produce higher quality maps with less processing power. Throughout the experiment, sets of $1$ to $12$ robots' local maps were randomly chosen and fused together using the approach described in Section~\ref{sec:algorithm}. Each experiment was repeated 24 times to control for between-run variation in the topic models produced by each robot.

The topic model hyperparameters were set using a Bayesian Optimization algorithm~\cite{Cully2018} to find the values that resulted in the highest map quality, evaluated using a third simulated reef environment which was similar to Environment \#1. The topic model vocabulary was the same one used in~\cite{Girdhar2019_ICRA}. %
All code, instructions, hyperparameters, and datasets required to reproduce these experiments, as well as instructions to generate similar test environments, are available in the Sunshine repository.%

\subsection{Evaluating Semantic Map Quality}

A useful fused semantic map $g(x)$ is one that is, at every location $x$, a \textit{good predictor of} the ground truth label $f(x)$ defined by the human operator. We measure this predictive strength using Adjusted Mutual Information (AMI)~\cite{Vinh2009}, a normalized variant of Mutual Information (MI). MI represents the number of bits of information contained in one random variable that describe another, and is defined for discrete random variables $U\in\mathcal{U},V\in\mathcal{V}$ as
\begin{equation}
    \text{MI}(U, V) = \sum_{u\in \mathcal{U}} \sum_{v\in \mathcal{V}} p(u,v)\log\left(\frac{p(u)p(v)}{p(u,v)}\right).
\end{equation}

Given the semantic maps $f$ and $g$ and a finite set of cells $\mathcal{B}$, we define the random variable $B$ as a cell chosen uniformly at random from $\mathcal{B}$, and thus define the random variables $Y_f=f(B)$ and $Y_g=g(B)$ as the robot team's semantic label and the ground truth semantic label for $B$, respectively. The joint probability of these random variables is:
\begin{equation}\label{eq:joint-prob}
    p(y_f, y_g) = \frac{1}{\vert\mathcal{B}\vert}\sum_{b\in\mathcal{B}} \mathds{1}_{f(b)=y_f \wedge g(b) = y_g},
\end{equation}
which gives the probabilities $p(y_f), p(y_g)$ through marginalization. Denoting the entropy of random variables $Y_f$ and $Y_g$ as $H(Y_f)$ and $H(Y_g)$, the AMI is computed as
\begin{equation} \hspace*{-0.05in}
    \text{AMI}(Y_f, Y_g) \!= \!\! \frac{\text{MI}(Y_f,Y_g) - \mathbb{E}[\text{MI}(Y_f, Y_g)]}{\max\{H(Y_f), H(Y_g)\} - \mathbb{E}[\text{MI}(Y_f,Y_g)]}.
\end{equation}

A perfect AMI score of 1 indicates that the robot team's semantic map contains all information required to reproduce the ground truth semantic map of the same area. Conversely, a score of $0$ indicates that the team's semantic map contains no more information about the corresponding part of the ground truth map than a randomly generated map is expected to. This is because the normalization subtracts out the expected mutual information $\mathbb{E}[\text{MI}(Y_f,Y_g)]$ between the labelings $Y_f$ and $Y_g$, computed according to~\cite{Vinh2009}. If a fused semantic map has a high AMI score with respect to the ground truth, then there is a consistent correspondence between each semantic label used by the robot team and each label used to produce the ground truth map. Thus, for high AMI scores, the human operator only needs to look at a few example images for any label in the fused map in order to determine that label's human-interpretable meaning.

\subsection{Baseline Comparisons}

The local maps were also fused using the ID-based matching and Hungarian matching approaches described in~\cite{Doherty2018} to get baseline performance metrics. ID-based matching assumes that every robot observed the same phenomena in the same order, and so the first topic learned by one robot corresponds to the first topic learned by every other robot, and likewise for the remaining topics. The Hungarian approach only assumes that every robot observed the same phenomena, and finds the maximum similarity permutation between the first robot's topics and each additional robot's. This is a \textit{sequential}, not multiway, matching approach because it compares topics belonging to a pair of robots at a time instead of considering all of the topics together. We are not aware of any previous baselines that used a multiway matching algorithm or did not assume that every robot observed the same phenomena.

Separate from the multi-robot experiments, a single robot was used to explore the same environments and independently build its own semantic map. It used the same topic model hyperparameters as the robots in the multi-agent experiment, but did not require any topic matching or map fusion. The single robot required $Nt$ seconds to explore the same area that $N$ robots explored in $t$ seconds; the quality of the map it produced after exploring the same area as the $N$ fused robots is reported in the results as ``Single Robot''.

\section{Results \& Discussion}
\label{sec:results}

\begin{figure}[t]
    \centering
    \subcaptionbox{\label{fig:gt-map-1}Ground truth segmentation.}{
    \includegraphics[width=.45\columnwidth]{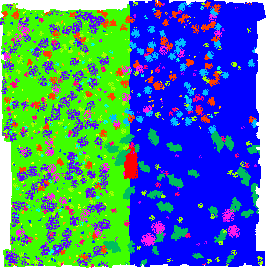}
    }
    \subcaptionbox{\textit{Ours} (CLEAR Based).}{
    \includegraphics[width=.45\columnwidth]{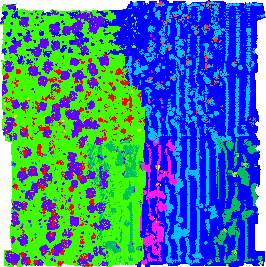}
    }
    \subcaptionbox{Hungarian matching.}{
    \includegraphics[width=.45\columnwidth]{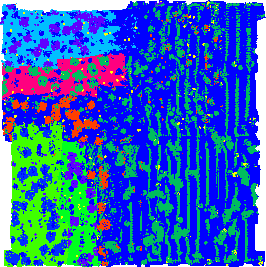}
    }
    \subcaptionbox{ID-Based matching.}{
    \includegraphics[width=.45\columnwidth]{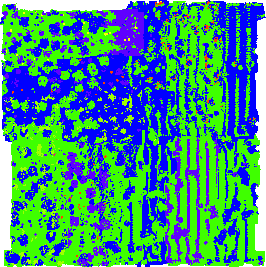}
    }
    \caption{Sample fused maps from each multi-robot matching approach with all 12 robots, alongside the ground truth segmentation, for Environment \#2. Note that each map has been manually colored (1 color per label) with the same palette to ease comparison. Our approach most accurately captures the variation in terrain and coral species present in each quadrant.}
    \label{fig:example-maps}
\end{figure}

Figure~\ref{fig:gt-ami} shows the performance of the proposed system using CLEAR, compared to using baseline matching solutions and to using a single robot. The performance was measured in terms of AMI of the fused map with the ground truth semantic map produced by AirSim. While the performance of the other matching algorithms declines as more robots are fused, the performance of the proposed matching solution \textit{increases} or stays steady. This happens because CLEAR leverages redundant edges in the noisy association graph, added by additional robots, to help compensate for incorrect edges. The number of incorrect edges at each vertex grows slower than the number of correct edges as topics are added, so our system is able to find a better solution when more robots' maps are fused. 

In the first test environment, Fig.~\ref{fig:gt-ami-1}, the fused map quality of the proposed approach goes from about 10\% lower than the semantic map produced by a single robot to about 10\% higher as more robots are fused. This is excellent performance considering that the robot team was able to map the entire environment in $1/12^\text{th}$ the amount of time.
Compared to the Hungarian matching approach, the proposed system achieves 23\% higher AMI scores on average; as shown in Fig.~\ref{fig:example-maps}, this is primarily because CLEAR is better suited to recognize when different robots have observed distinct phenomena. In the second test environment, Fig.~\ref{fig:gt-ami-2}, this difference was magnified as there was very little in common between what any pair of robots observed. Table~\ref{tab:results} summarizes numerical results for the map quality after fusing all 12 local maps together with each matching algorithm and for various similarity and distance metrics. The other figures shown used Cosine similarity for CLEAR matching and Euclidean (L2) distance as the Hungarian cost metric.

\begin{table}
\caption{Semantic Mapping Performance with 12 Robots.\label{tab:results}}
\begin{tabular*}{\columnwidth}{@{}llcc@{}}
\toprule
\multicolumn{2}{c}{}                                  & \multicolumn{2}{c}{\textsc{Mean AMI Score (Std. Dev.)}} \\ 
 \multicolumn{1}{c}{Matching Alg.} & \multicolumn{1}{c}{Metric} & \multicolumn{1}{c}{Env. \#1} & \multicolumn{1}{c}{Env. \#2}\\
\midrule
ID-Based &       N/A                   & 0.117 (0.035)                                                               &          0.078 (0.016)                                                      \\
Hungarian              & L1 Distance                &                                    0.297 (0.006)                           &                                     0.216 (0.004)                           \\
                    & L2 Distance                  &                         0.313 (0.016)                                      &      0.253 (0.039)                                                          \\
                    & Cosine Distance              &      0.304 (0.011)                                                         &     0.203 (0.015)                                                           \\
CLEAR                  & TO Similarity                &                         0.250 (0.002)                                      &                        0.341 (0.003)                                        \\
                     & Cosine Similarity            &     \textbf{0.384} (0.006)                                                          &             \textbf{0.406} (0.007)                                                   \\\midrule
\multicolumn{2}{l}{Single Robot (No Matching)} & 0.344 (0.026) & 0.382 (0.024) \\\bottomrule
\end{tabular*}
\end{table}

\begin{figure}[t]
    \centering
    \includegraphics[width=\columnwidth]{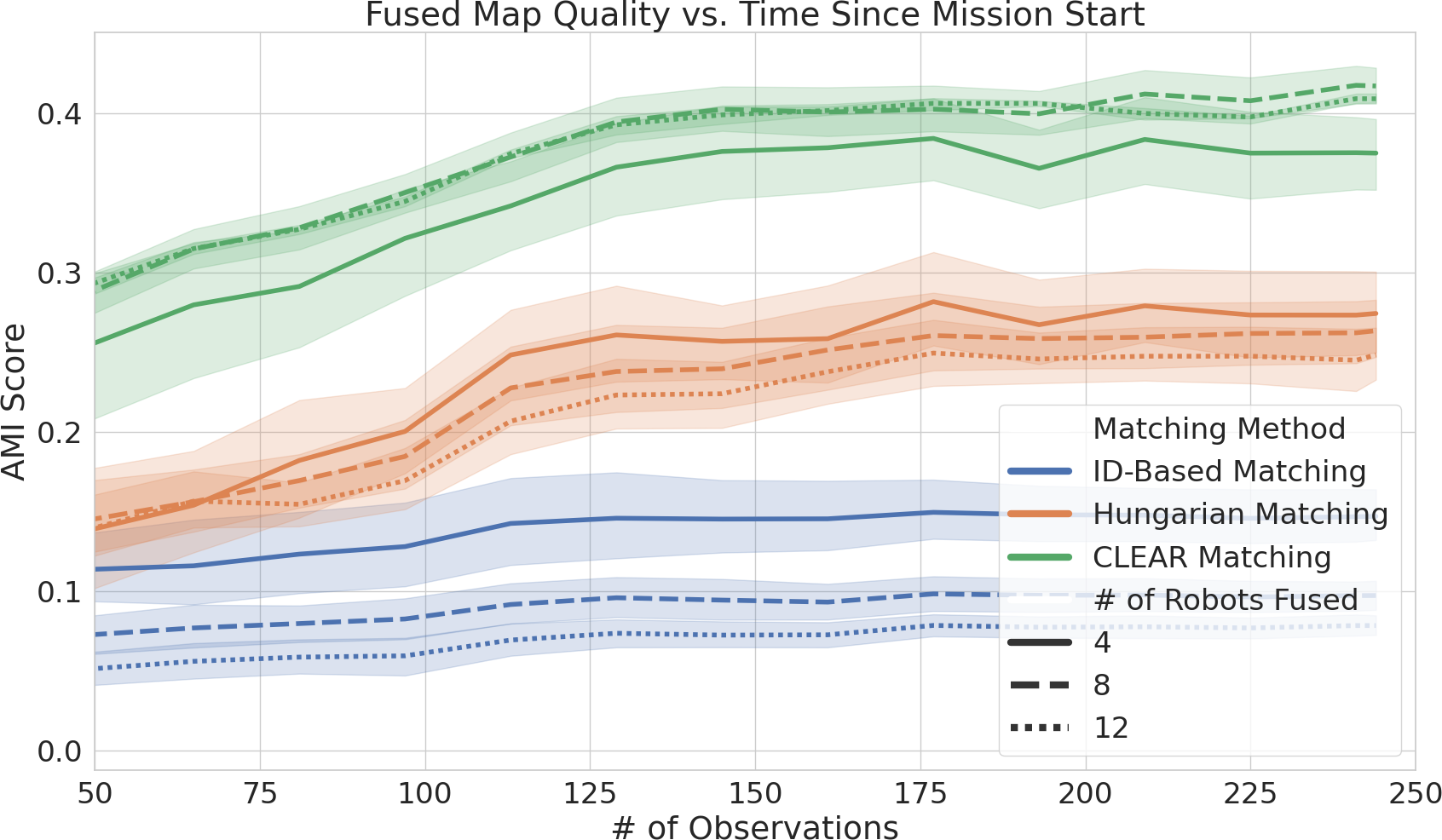}
    \caption{The fused map quality varies throughout each experiment; shown here is how the fused map quality (AMI) changes as the robots explore environment \#2. The increase in performance across all methods is caused by each robot's topic model improving over time with more data.}
    \label{fig:gt-ami-obs}
\end{figure}

As seen in Fig.~\ref{fig:gt-ami-obs}, as the team explores environment \#2 the fused map quality is mostly constant after each robot has collected 125 images, i.e., covered half of its assigned area. This suggests that environment \#2 would be most efficiently explored using 24 robots; in general, the optimal number will depend on the size and complexity of the environment. %

The results presented are expected to generalize well to real world environments. Previous versions of BNP-ROST have demonstrated the ability to produce useful semantic maps of real-world environments while running on real robot hardware~\cite{Girdhar2019_ICRA}. Furthermore, the compressed semantic maps and topic descriptors shared between robots are very small (typically around $10$ to $100$ kB) so they can be transmitted between robots in even severely bandwidth-constrained environments, like the deep sea~\cite{Girdhar2019_ICRA,Kaeli2014}.

\section{Conclusions}

We have presented a novel multi-robot distributed semantic mapping system that produces accurate semantic maps even when fusing maps from \textit{many} robots and when each robot is building its unsupervised semantic model online with \textit{no pre-training}. The proposed topic matching approach results in 20-60\% higher map quality than pairwise Hungarian matching, with the largest gains in mapping complex and diverse environments, while also using less communication bandwidth than the previous state-of-the-art~\cite{Doherty2018}. The fused maps are suitable for the human operator to use for mission summarization and informative path planning. We find that the fused maps approximate the quality of the best single-robot maps, hence further performance increases will likely come from improving the STM-based online semantic mapping component. The presented system for accurate topic matching over low-bandwidth enables novel multi-robot distributed autonomous exploration capabilities, such as cooperative-adaptive path planning and distributed reward learning, which will be explored in future work.%

\clearpage
\balance %

\bibliographystyle{IEEEtran}
\ifthenelse{\boolean{arxiv}}{

}{
\bibliography{IEEEabrv,stewart}
}

\end{document}